\pdfoutput=1
 
\documentclass[11pt]{article}

\usepackage{authblk}

\usepackage[]{naacl2021}

\usepackage{times}
\usepackage{latexsym}

\usepackage[T1]{fontenc}

\usepackage[utf8]{inputenc}

\usepackage{microtype}

\newcommand{\CoolName}{MTAG\xspace}

\usepackage{tikz}

\newcommand{\Simley}[1]{%
\begin{tikzpicture}[scale=0.11]
    \newcommand*{\SmileyRadius}{1.0}%
    \draw [fill=brown!10] (0,0) circle (\SmileyRadius)
        ;  

    \pgfmathsetmacro{\eyeX}{0.5*\SmileyRadius*cos(30)}
    \pgfmathsetmacro{\eyeY}{0.5*\SmileyRadius*sin(30)}
    \draw [fill=cyan,draw=none] (\eyeX,\eyeY) circle (0.15cm);
    \draw [fill=cyan,draw=none] (-\eyeX,\eyeY) circle (0.15cm);

    \pgfmathsetmacro{\xScale}{2*\eyeX/180}
    \pgfmathsetmacro{\yScale}{1.0*\eyeY}
    \draw[color=red, domain=-\eyeX:\eyeX]   
        plot ({\x},{
            -0.1+#1*0.15 
            -#1*1.75*\yScale*(sin((\x+\eyeX)/\xScale))-\eyeY});
\end{tikzpicture}%
}%

\usepackage{xcolor}
\usepackage{amsmath}
\usepackage{booktabs} 
\DeclareMathAlphabet\mathbfcal{OMS}{cmsy}{b}{n}

\usepackage[linesnumbered,ruled,vlined]{algorithm2e}

\SetCommentSty{mycommfont}

\DeclareMathAlphabet\mathbfcal{OMS}{cmsy}{b}{n}

\usepackage{lipsum}

\usepackage{multirow}
\usepackage{multicol}

\DeclareMathOperator*{\concat}{concat}

\usepackage{breqn}

\usepackage{graphicx}

\usepackage{subcaption}

\title{\CoolName: Modal-Temporal Attention Graph for\\Unaligned Human Multimodal Language Sequences}

\newcommand*\samethanks[1][\value{footnote}]{\footnotemark[#1]}

\author[ ]{Jianing Yang\textsuperscript{\rm 1}\thanks{\ \ Equal contribution}}
\author[ ]{Yongxin Wang\textsuperscript{\rm 1}\samethanks}
\author[1]{Ruitao Yi}
\author[1]{Yuying Zhu}
\author[1]{Azaan Rehman}
\author[1]{\\Amir Zadeh}
\author[2]{Soujanya Poria}
\author[1]{Louis-Philippe Morency}
\affil[1]{Carnegie Mellon University}
\affil[2]{Singapore University of Technology and Design}
\affil[ ]{\small{\{jianing3, yongxinw, ruitaoy, yuyingz, arehman, abagherz\}@cs.cmu.edu, sporia@sutd.edu.sg, morency@cs.cmu.edu}}

\begin{document}
\maketitle

\begin{abstract}
Human communication is multimodal in nature; it is through multiple modalities such as language, voice, and facial expressions, that opinions and emotions are expressed. Data in this domain exhibits complex multi-relational and temporal interactions. Learning from this data is a fundamentally challenging research problem. In this paper, we propose Modal-Temporal Attention Graph (\CoolName). \CoolName is an interpretable graph-based neural model that provides a suitable framework for analyzing multimodal sequential data. We first introduce a procedure to convert unaligned multimodal sequence data into a graph with heterogeneous nodes and edges that captures the rich interactions across modalities and through time. Then, a novel graph fusion operation, called \CoolName fusion, along with a dynamic pruning and read-out technique, is designed to efficiently process this modal-temporal graph and capture various interactions. By learning to focus only on the important interactions within the graph, \CoolName achieves state-of-the-art performance on multimodal sentiment analysis and emotion recognition benchmarks, while utilizing significantly fewer model parameters.\footnote{Code is available at \url{https://github.com/jedyang97/MTAG}.}
\end{abstract}

\section{Introduction}
\begin{figure}[ht]
    \centering
    \includegraphics[width=\linewidth]{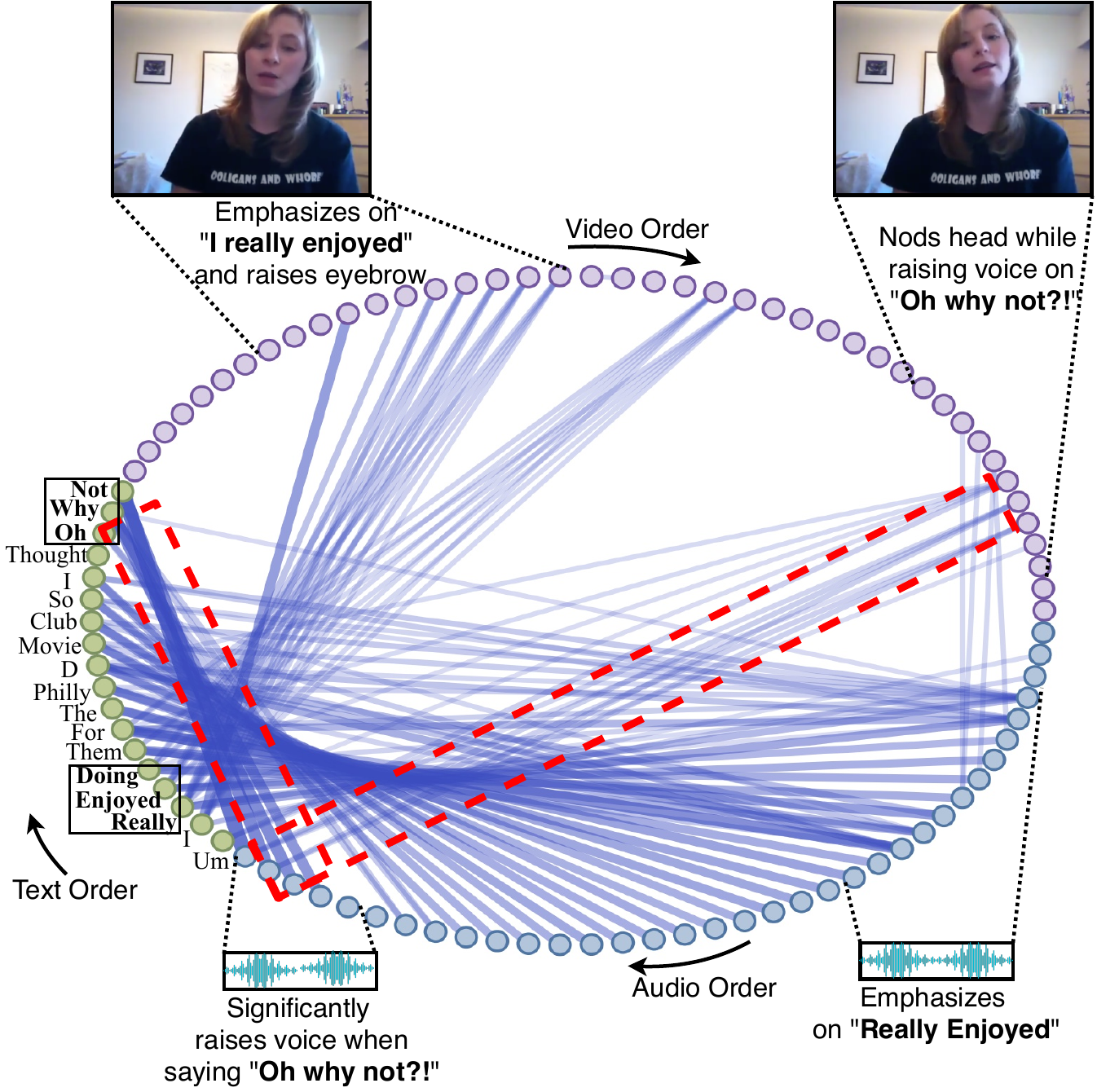}
    \caption{Example visualization of tri-modal Modal-Temporal Attention learned by our proposed model. Each circle represents a node from video/text/audio modalities, and the blue lines denote the learned attention weights (\textit{i.e.} the thicker and darker a blue line is, the larger the attention weight). We observe high intensities between semantically correlated graph entities, such as "Really Enjoy" and the raise in eyebrow, which indicate positive sentiment. Note that our graph-based model learns multimodal interactions without prior alignment, and captures diverse types of interactions across multiple modalities all the same time. Edge types are not shown for visual clarity.}
    \label{fig:tri-modal-vis}
\end{figure}

With recent advances in machine learning research, analysis of multimodal sequential data has become increasingly prominent. At the core of modeling this form of data, there are the fundamental research challenges of \textit{fusion} and \textit{alignment}. Fusion is the process of blending information from multiple modalities. It is usually preceded by alignment, which is the process of finding temporal relations between the modalities. An important research area that exhibits this form of data is multimodal language analysis, where sequential modalities of language, vision, and acoustic are present. These three modalities carry the communicative information and interact with each other through time; e.g. positive word at the beginning of an utterance may be the cause of a smile at the end. 
\begin{figure*}[t]
    \centering
    \includegraphics[width=0.92\textwidth]{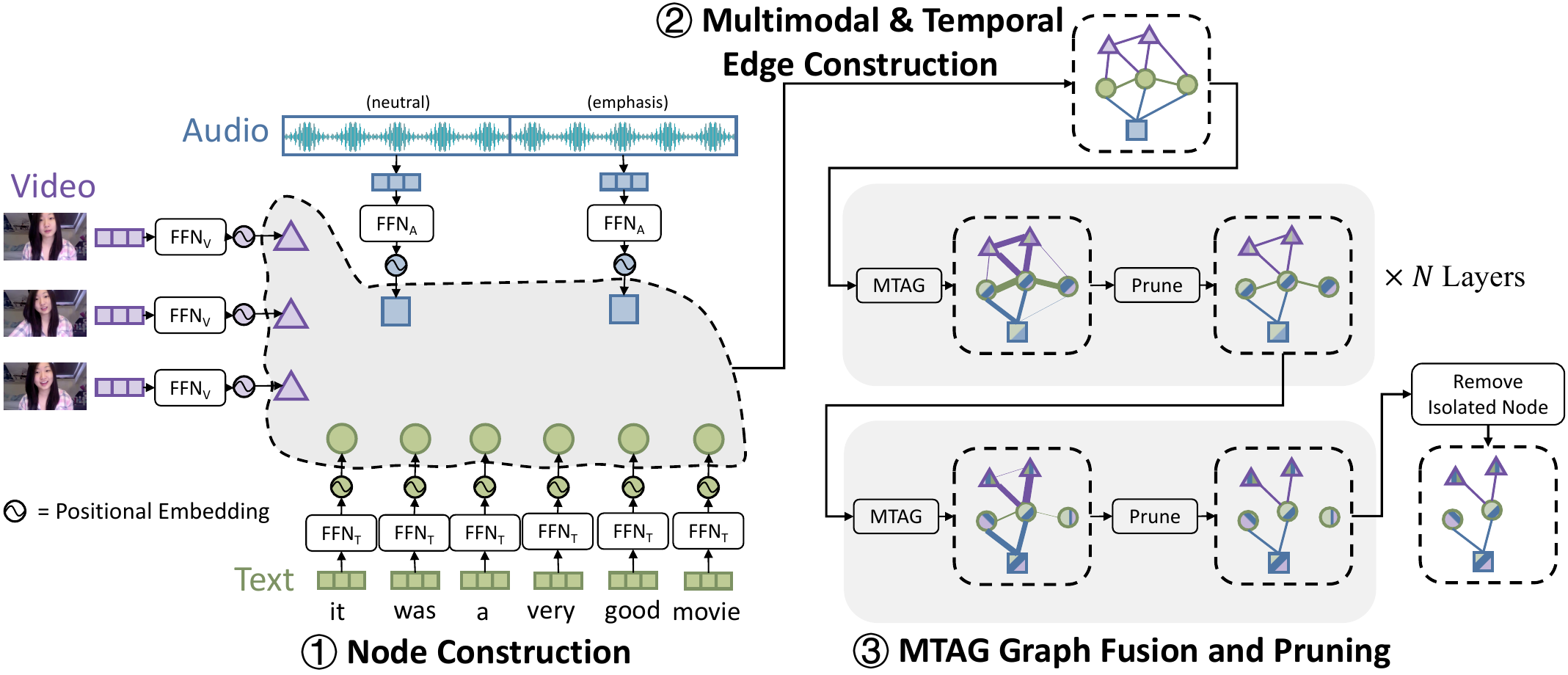}
    \caption{The 3-stage \CoolName framework: Node Construction, Edge Construction and Fusion+Pruning. \textbf{[Node Construction]} Each modality's features are first passed through a distinct Feed-Forward-Network to be mapped into the same embedding size. Then, a positional embedding is added to each transformed feature based on its position in its own modality, so that temporal information are encoded. The features are now nodes in the graph. \textbf{[Edge Construction]} We then apply an algorithm to construct edges among these nodes by appropriately indexing each edge with a modal type and a temporal type. \textbf{[Fusion+Pruning]} Finally, we pass the graph into the \CoolName module to learn interactions across modality and time. The output graph with updated node embeddings and pruned edges can be passed to downstream modules, e.g. a Multi-layer Perceptron, to complete specific tasks such as regression or classification.}
    \label{fig:system}
\end{figure*}
When analyzing such multimodal sequential data, it is crucial to build models that perform both fusion and alignment accurately and efficiently by a) aligning arbitrarily distributed asynchronous modalities in an interpretable manner, b) efficiently accounting for short and long-range dependencies, c) explicitly modeling the inter-modal interactions between the modalities while simultaneously accounting for intra-modal dynamics.

In this paper, we propose \CoolName (Modal-Temporal Attention Graph). \CoolName is capable of both fusion and alignment of asynchronously distributed multimodal sequential data. Modalities do not need to be pre-aligned, nor do they need to follow similar sampling rate. \CoolName can capture interactions of various types across any number of modalities all at once, comparing to previous methods that model bi-modal interactions at a time \cite{tsai-etal-2019-multimodal}. At its core, \CoolName utilizes an efficient trimodal-temporal graph fusion operation. Coupled with our proposed dynamic pruning technique, \CoolName learns a parameter-efficient and interpretable graph. In our experiments, we use two unaligned multimodal emotion recognition and sentiment analysis benchmarks: IEMOCAP \cite{busso2008iemocap} and CMU-MOSI \cite{zadeh2016multimodal}. The proposed \CoolName model achieves state-of-the-art performance with far fewer parameters. Subsequently, we visualize the learned relations between modalities and explore the underlying dynamics of multimodal language data. Our model incorporates all three modalities in both alignment and fusion, a fact that is also substantiated in our ablation studies.

\section{Related Works}

\paragraph{Human Multimodal Language Analysis}
Analyzing human multimodal language involves learning from data across multiple heterogeneous sources that are often asynchronous, i.e. language, visual, and acoustic modalities that each uses a different sampling rate. Earlier works assumed multimodal sequences are aligned based on word boundaries \cite{lazaridou2015combining,ngiam2011multimodal,gu-etal-2018-multimodal,dumpala2019audio,pham2019found} and applied fusion methods for aligned sequences.
To date, modeling \emph{unaligned} multimodal language sequences remains understudied, except for \cite{tsai-etal-2019-multimodal,Khare_2020,zheng2020crossmodality}, which used cross-modal Transformers to model unaligned multimodal language sequences. However, the cross-modal Transformer module is a bi-modal operation that only account for two modalities' input at a time. In \citet{tsai-etal-2019-multimodal}, the authors used multiple cross-modal Transformers and applies late fusion to obtain tri-modal features, resulting in a large amount of parameters needed to retain original modality information. Other works that also used cross-modal Transformer architecture for  include \citet{10.1145/3394171.3413690,Siriwardhana2020JointlyF}. In contrast to the existing works, our proposed graph method, with very small amount of model parameters, can aggregate information from \emph{multiple} (more than 2) modalities at early stage by building edges between the corresponding modalities, allowing richer and more complex representation of the interactions to be learned. 
\paragraph{Graph Neural Networks}
Graph Neural Network (GNN) was introduced in \cite{gori2005new,scarselli2008graph} with an attempt to extend deep neural networks to handle graph-structured data. Since then, there has been an increasing research interest on generalizing deep neural network's operations such as convolution \cite{kipf2016semi, schlichtkrull2017modeling, hamilton2017inductive}, recurrence \citep{nicolicioiu2019recurrent}, and attention \cite{velickovic2018graph} to graph.

Recently, several \emph{heterogeneous} GNN methods \cite{wang_heterogeneous_2019, wei_mmgcn_2019,shi2016survey} have been proposed. The heterogeneous nodes referred in these works consist of uni-modal views of \emph{multiple} data generating sources (such as movie metadata node, audience metadata node, etc.), whereas in our case the graph nodes represent \emph{multimodal} views of a \textit{single} data generating source (visual, acoustic, textual nodes from a single speaking person). In the NLP domain, multimodal GNN methods \cite{khademi-2020-multimodal,yin-etal-2020-novel} on tasks such as Visual Question Answering and Machine Translation. However, these settings still differ from ours because they focused on static images and short text which, unlike the multimodal video data in our case, do not exhibit long-term temporal dependencies across modalities. 

Based on these findings, we discovered there has been little research using graph-based methods for modeling \emph{unaligned, multimodal} language sequences, which includes video, audio and text. In this paper, we demonstrate our proposed \CoolName method can effectively model such unaligned, multimodal sequential data.
\section{\CoolName}

\begin{table}[h]
\resizebox{\linewidth}{!}{
	\begin{tabular}{ccl}
		\toprule
		Notation&Explanation\\
		\midrule
		
		
		$v_{i}$ & Node $i$ \\
		$e_{ij}$ & Edge from $v_{i}$ to $v_{j}$ \\
		$\mathbfcal{N}_i$ & Neighbor nodes incident into $v_{i}$\\
		${\mathbf{x}_i}$& Initial node feature for $v_{i}$ \\
		${\mathbf{x'}_i}$& Transformed node feature for $v_{i}$ \\
		
		${\pi}_i$ & Node type for $v_i$\\
		${\phi}_{ij}$ & Edge modality type for $e_{ij}$ \\
		${\tau}_{ij}$ & Edge temporal type for $e_{ij}$ \\
		
		$\mathbf{M}_{\pi_i}$ & Node type specific transformation matrix \\

		$\mathbf{a}^{\phi_{ij}, \tau_{ij}}$ & Edge type specific learnable attention vector \\
		$\beta_{i,j}$ & Raw attention score of node pair ($v_i$,$v_j$) \\
		\multirow{2}{*}{$\alpha_{i,j}$} &  Attention weight of node pair ($v_i$,$v_j$),\\ & normalized over $\mathbfcal{N}_i$ \\
        ${\mathbf{z}_i}$& Node output feature for $v_i$ \\
        $k$ & Prune percentage \\
        $h$ & Index of multi-head attention head \\
        $H$ & Number of total attention heads\\
		\bottomrule
		
	\end{tabular}
}
	\caption{Terminologies used in this paper.}
	\label{tab:terminologies}
\end{table}

In this section, we describe our proposed framework: Modal Temporal Attention Graph (\CoolName) for unaligned multimodal language sequences. We describe how we formulate the multimodal data into a graph $\mathbfcal{G}(\mathbfcal{V}, \mathbfcal{E})$, and the \CoolName fusion operation that operates on $\mathbfcal{G}$. In essence, our graph formulation by design alleviates the need for any hard alignments, and combined with \CoolName fusion, allows nodes from one modality to interact freely with nodes from all other modalities at the same time, breaking the limitation of only modeling pairwise modality interactions in previous works.
Figure \ref{fig:system} gives a high-level overview of the framework.

\subsection{Node Construction}
As illustrated in Figure \ref{fig:system}, each modality's input feature vectors are first passed through a modality-specific Feed-Forward-Network. This allows feature embeddings from different modalities to be transformed into the same dimension. 
A positional embedding (details in Appendix \ref{sec:appendix}) is then added (separately for each modality) to each embedding to encode temporal information. The output of this operation becomes a node $v_i$ in the graph. Each node is marked with a modality identifier $\pi_i$, where $\pi_i \in \{\text{Audio, Video, Text}\}$ in our case.


\subsection{Edge Construction}
In this section, we describe our design of modality edges and temporal edges. For a given node of a particular modality, its interactions with nodes from different modalities should be considered differently. For example, given a Video node, its interaction with an Audio node should be different from that with a Text node. In addition, the temporal order of the nodes also plays a key role in multimodal analysis \cite{poria-etal-2017-context}. For example, a transition from a frown to a smile (\Simley{-.8} $\rightarrow$ \Simley{0} $\rightarrow$\Simley{.8}) may imply a positive sentiment, whereas a transition from a smile to a frown (\Simley{.8} $\rightarrow$ \Simley{0} $\rightarrow$\Simley{-.8}) may imply a negative sentiment. Therefore, interactions between nodes that appear in different temporal orders should also be considered differently. In GNNs, the edges define how node features are aggregated within a graph. In order to encapsulate the diverse types of node interactions, we assign edge types to each edge so that information can be aggregated differently on different types of edges. By indexing edges with edge types, different modal and temporal interactions between nodes can be addressed separately. 

\noindent{\textbf{Multimodal Edges.} 
As we make no assumption about prior alignment of the modalities, the graph is initialized to be a fully connected graph. We use $e_{ij}$ to represent an edge from $v_i$ to $v_j$. We assign $e_{ij}$ with a modality type identifier $\phi_{ij} = (\pi_i \rightarrow \pi_j)$. For example, an edge pointing from a Video node to a Text node will be marked with type $\phi_{ij} = (\text{Video} \rightarrow \text{Text})$.}

\noindent\textbf{Temporal Edges.} In addition to $\phi_{ij}$, we also assign a temporal label $\tau_{ij}$ to each $e_{ij}$. Depending on the temporal order of $v_i$ and $v_j$ connected by $e_{ij}$, we determine the value of $\tau_{ij}$ to be either of  $\{\text{past, present, future}\}$. For nodes from the same modality, the temporal orders can be easily determined by comparing their order of occurrences. To determine the temporal orders for nodes across different modalities, we first roughly align the two modalities with our 
pseudo-alignment. Then the temporal order can be simply read out.

\begin{figure}[t]
    \centering
    \includegraphics[width=0.8\linewidth]{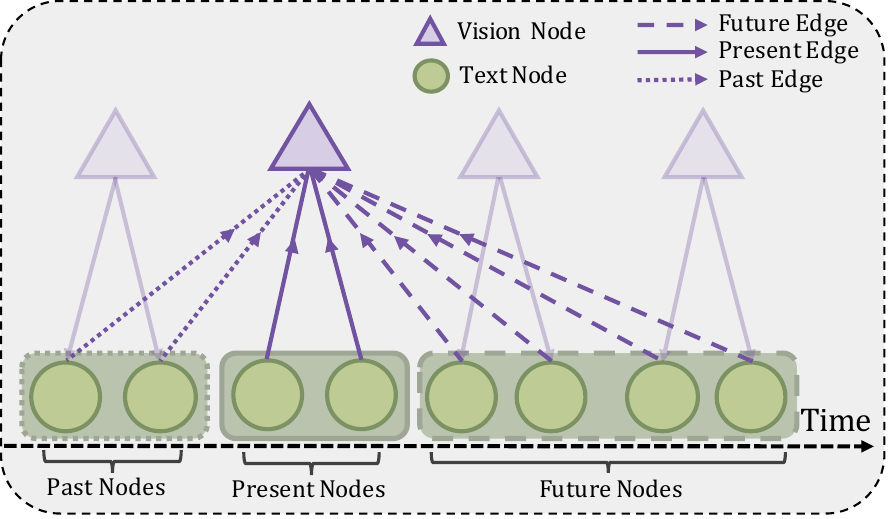}
    \caption{An example of the pseudo-alignment between two unaligned sequences. We first align the longer sequence to the shorter one as uniformly as possible. Then the aligned nodes from the longer sequence becomes the $\mathbfcal{V}_{i, present}$ for node $v_i$ in the shorter sequence. $\mathbfcal{V}_{i, past}$ and $\mathbfcal{V}_{i, future}$ can then be determined accordingly.}
    \label{fig:alignment}
\end{figure}

\noindent\textbf{Pseudo-Alignment.} As mentioned above, it is simple to determine the temporal edge types for nodes in a single modality. However, there is no clear definition of ``earlier" or ``later" across two modalities, due to the unaligned nature of our input sequences. To this end, we introduce the pseudo-alignment heuristic that coarsely defines the past, present and future connections between nodes across two modalities. Given a node $v_i$ from one modality $\pi_i$, our pseudo-alignment first determines a set of nodes $\mathbfcal{V}_{i, present}$ in the other modality that can be aligned to $v_i$ and considered as ``present". All nodes in the other modality that exists after $\mathbfcal{V}_{i, present}$ are considered ``future" $\mathbfcal{V}_{i, future}$, and all those before are considered $\mathbfcal{V}_{i, past}$. Once the coarse temporal order is established, the cross-modal temporal edge types can be easily determined. Figure \ref{fig:alignment} shows an example of such pseudo-alignment, and more details regarding the calculations can be found in Appendix \ref{sec:appendix_pseudo_align}.



\subsection{Fusion and Pruning}

\subsubsection{MTAG Fusion}
\begin{figure}
    \centering
    \includegraphics[width=.8\linewidth]{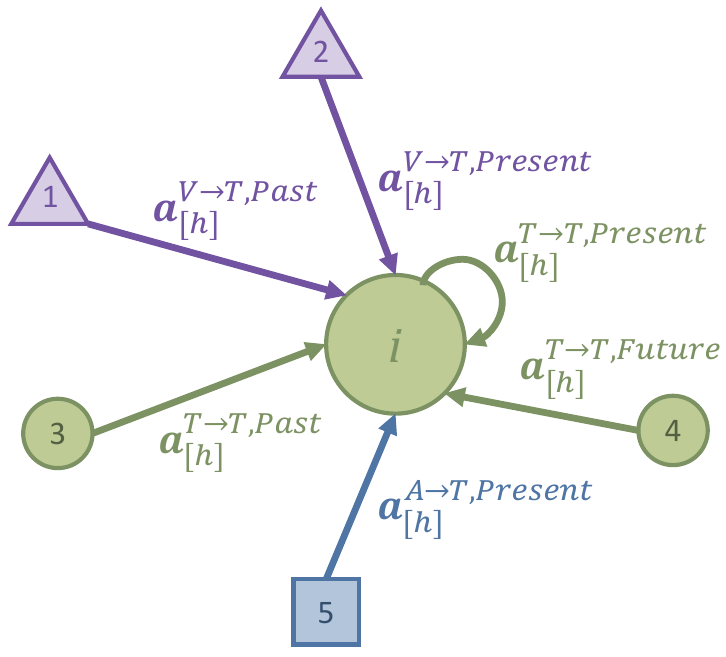}
    \caption{Visualization of the \CoolName operation around a single node. The text on each edge indicates which attention vector is used for that edge. Purple triangle represents a video node, green circle represents a text node and blue square represents an audio node.}
    \label{fig:mtgat}
\end{figure}

With our formulation of the graph, we design the \CoolName fusion operation that can digest our graph data with various node and edge types, and thus model the modal-temporal interactions.
An algorithm of our method is shown in Algorithm \ref{algo:mtgat} and a visual illustration is given in Figure \ref{fig:mtgat}. Specifically, for each neighbor node $v_j$ that has an edge incident into a center node $v_i$, we compute a raw attention score $\beta_{[h],i,j}$ based on that edge's modality and temporal type:
\begin{align}
        \beta_{[h],i,j} = \text{LeakyRelu}(\mathbf{a}_{[h]}^{\phi_{ji}, \tau_{ji}} \cdot [\mathbf{x}_i'\Vert \mathbf{x}_j'])
\end{align}
where $[\cdot||\cdot]$ denotes the concatenation of two column vectors into one long column vector. The $[h]$ index is used to distinguish which multi-head attention head is being used. Note that $\mathbf{a}_{[h]}^{\phi_{ji},\tau_{ji}}$ depends on both the modality and temporal edge types of $e_{ji}$. This results in 27 edge types (9 types of modality interaction $\times$ 3 types of temporal interaction).

We normalize the raw attention scores over all neighbor nodes $v_j$ with Softmax so that the normalized attention weight sums to 1 to preserve the scale of the node features in the graph.
\begin{equation}
\alpha_{[h],i,j}
=\frac{\exp \bigl( \beta_{[h],i,j} \bigr)}{\sum_{k\in \mathbfcal{N}_i} \exp \bigl( \beta_{[h],i,k} \bigr)}
\end{equation}

Then, we perform node feature aggregation for each node $v_i$ following:
\begin{equation}
\mathbf{z}_i = \concat\limits_{h=1}^H(\sum_{j\in \mathbfcal{N}_i} \alpha_{[h],i,j} \mathbf{x}_j')
\end{equation}
where $\mathcal{N}_i$ defines the neighbors of $v_i$ and hyperparameter $H$ is the number of total attention heads. $\mathbf{z}_i$ now becomes the new node embedding for node $v_i$. After aggregation, $v_i$ transformed from a node with unimodal information into a node encoding the diverse modal-temporal interactions between $v_i$ and its neighbors (illustrated by the mixing of colors of the nodes in Figure \ref{fig:system}).

\begin{algorithm}[t]
\DontPrintSemicolon
\SetNoFillComment
\SetKwInOut{Input}{Input}
\SetKwInOut{Output}{Output}
Feature transformation $\mathbf{x}'_i \gets \mathbf{M}_{\pi_i} \mathbf{x}_i, \forall i$ \\

\For(){$h = 1...H$}{
    \For(){$j \in \mathbfcal{N}_i \cup i$}{
    
    calculate raw attention score using modality- and temporal-edge-type specific parameters: $\beta_{[h],i,j} = \text{LeakyRelu}(\mathbf{a}_{[h]}^{\phi_{ji}, \tau_{ji}} \cdot [\mathbf{x}_i'\Vert \mathbf{x}_j'])$

    }
    normalize raw attention scores over $\mathbfcal{N}_i  \cup i$  to get attention weight $\alpha_{[h],i,j}$

}
calculate node output feature $\mathbf{z}_i = \concat\limits_{h=1}^H(\sum_{j\in \mathbfcal{N}_i  \cup i} \alpha_{[h],i,j} \mathbf{x}_j')$

calculate average attention weight across all heads $\overline{\alpha}_{i,j} = \frac{1}{H}\sum_{h=1}^H(\alpha_{[h],i,j})$\\
sort $\overline{\alpha}_{i,j}$ and delete the edges with the smallest $k\%$ average attention weight from $\mathbfcal{N}_i  \cup i$, obtaining $\mathbfcal{N}_i'$

\Return $\mathbf{z}_i$, $\mathbfcal{N}_i' \ \   \forall i$

\caption{\CoolName with edge pruning}
\label{algo:mtgat}
\end{algorithm}

We desgined the operation to have $H$ multi-head attention heads because the heterogeneous input data of the multimodal graph could be of different scales, making the variance of the data high. Adding multi-head attention could help stabilize the behavior of the operation.

\subsubsection{Dynamic Edge Pruning}

Our graph formulation models interactions for all 27 edge types. This design results in a very large number of edges in the graph, making the computation graph difficult to fit into GPU memories.  More importantly, when there are so many edges, it is hard to avoid some of these edges from inducing spurious correlations and distracting the model from focusing on the truly important interactions \citep{Lee2019SelfAttentionGP, knyazev2019understanding}. To address these challenges, we propose to dynamically prune edges as the model learns the graph. Specifically, after each layer of \CoolName, we have the attention weight $\alpha_{[h],i,j}$ for each attention head $h$ and for each edge $e_{ij}$. We take the average of the attention weights over the attention heads:
\begin{equation}
\overline{\alpha}_{ij} = \frac{1}{H}\sum_{h=1}^H(\alpha_{[h],i,j})
\end{equation}
Then, we sort $\overline{\alpha}_{ij}$ and delete $k\%$ edges with the smallest attention weights, where $k$ is a hyperparameter. These deleted edges will no longer be calculated in the next \CoolName fusion layer. Our ablation study in Section \ref{sec:albation_pruning} empirically verifies the effectiveness of this approach by comparing to no pruning and random pruning.

\subsection{Graph Readout}

At the end of the \CoolName fusion process, we need to read out information scattered in the nodes into a single vector so that we can pass it through a classification head. Recall that the pruning process drops edges in the graph. If all edges incident into a node have been dropped, then it means that node was not updated based on its neighbors. In that case, we simply ignore that node in the readout process.
\begin{equation}
    \resizebox{1.0\linewidth}{!}{$\mathbfcal{V'} = \{v_i \ |\ v_i \in \mathbfcal{V} \text{ and } \text{count\_incident\_edge}(v_i) > 0\}$}
\end{equation}

We readout the graph by averaging all the surviving nodes' output features into one vector. This vector is then passed to a 3-layer Multi-Layer-Perceptron (MLP) to make the final prediction.
\section{Experiments}
\begin{table}[t]
\begin{center}
\setlength\tabcolsep{5.9pt}
\fontsize{9.5}{10}\selectfont
\renewcommand{\arraystretch}{1.2}
\begin{tabular}{c c c c c}
\toprule
\textbf{Model} \textbackslash \ \textbf{Emotion} & \textbf{Happy} & \textbf{Sad} & \textbf{Angry} & \textbf{Neutral} \\
\toprule
\multicolumn{5}{c}{\bf (Unaligned) IEMOCAP Emotions.}               \\ 
\toprule
CTC + EF-LSTM  & 75.7 & 70.5 & 67.1 & 57.4 \\ 
LF-LSTM  & 71.8 & 70.4  & 67.9 & 56.2 \\ 
CTC + RAVEN & 76.8  & 65.6  & 64.1 & 59.5\\ 
CTC + MCTN  & 77.5 & 71.7  & 65.6 &  49.3 \\ 
MulT  & 81.9 & 74.1 & 70.2  & 59.7 \\ 
\midrule
\CoolName (ours)  & \bf 86.0 & \bf 79.9  & \bf 76.7 &  \bf 64.1 \\  
\bottomrule
\end{tabular}
\caption{F1 Scores on unaligned IEMOCAP. Higher is better.}
\label{table:iemocap}
\end{center}
\end{table}

We empirically evaluate \CoolName model on two datasets: IEMOCAP \cite{busso2008iemocap} and CMU-MOSI \cite{zadeh2016multimodal}; both are well-known datasets used by prior works \citep{liang2018multimodal, pham2019found, tsai2019learning,  tsai-etal-2019-multimodal} to benchmark multimodal emotion recognition and sentiment analysis.

\subsection{Dataset and Metrics}
\paragraph{IEMOCAP} IEMOCAP is a multimodal emotion recognition dataset consisting of 10K videos.
The task we chose is the 4-way multilabel emotion classification, classifying into happy, sad, angry and neutral.
For train split, the positive/negative label ratio for each emotion is 954:1763, 338:2379, 690:2027 and 735:1982. For the test split, the ratio is 383:555, 135:803, 193:745 and 227:711. Due to this unbalanced distribution of the the labels, we use F1 score as a better metric for comparison.
\paragraph{CMU-MOSI} CMU Multimodal Opinion Sentiment Intensity is a multimodal sentiment analysis dataset with 2,199 movie review video clips. Each video clip is labeled with real-valued sentiment score within $[-3, +3]$, with $+3$ being a very positive sentiment and $-3$ a very negative one. Following previous works \cite{tsai-etal-2019-multimodal}, we report five metrics: 7-class classification accuracy ($\text{Acc}_7$), binary classification accuracy ($\text{Acc}_2$, positive/negative sentiments), F1 score, Mean Absolute Error (MAE) and the correlation of the model's prediction with human. 

We follow prior works \cite{tsai-etal-2019-multimodal} to evaluate on both of the above unaligned datasets, where original audio and video features are used, resulting in variable sequence lengths across modalities. For both datasets, the multimodal features are extracted from the textual (GloVe word embeddings \cite{pennington2014glove}), visual (Facet \cite{imotion}), and acoustic (COVAREP \cite{degottex2014covarep}) data modalities. 
\begin{table}[t]
\renewcommand\arraystretch{1.3}
\centering
\resizebox{\linewidth}{!}{
\begin{tabular}{c c c c c c} 
\toprule
\textbf{Model} \textbackslash \ \textbf{Metirc}       &  \textbf{$\text{Acc}_7\uparrow$} & \textbf{$\text{Acc}_2\uparrow$} & \textbf{$\text{F1}\uparrow$} & \textbf{$\text{MAE}\downarrow$} & \textbf{$\text{Corr}\uparrow$} \\ 
\toprule
\multicolumn{6}{c}{\bf(Unaligned) CMU-MOSI Sentiment}                  \\ \toprule
CTC+EF-LSTM & 31.0      & 73.6      & 74.5   & 1.078   & 0.542     \\
LF-LSTM       & 33.7      & 77.6      & 77.8   & 0.988   & 0.624     \\
CTC+MCTN    & 32.7      & 75.9      & 76.4   & 0.991   & 0.613     \\
CTC+RAVEN   & 31.7      & 72.7      & 73.1   & 1.076   & 0.544     \\ 
MulT   & \textbf{39.1}      & 81.1      &  81.0   & 0.889 & 0.686     \\ 
\midrule
\CoolName (ours)  & 38.9     &     \textbf{82.3}      &  \textbf{82.1}      & \textbf{0.866}        &   \textbf{0.722}        \\

\bottomrule
\end{tabular}
}
\caption{Results on unaligned CMU-MOSI.  $\uparrow$ means higher is better and $\downarrow$ means lower is better.}
\label{table:mosi}
\end{table}
\begin{table}[t]
\centering
\renewcommand\arraystretch{1.4}
\begin{tabular}{ c | c}
\hline
\textbf{Model} & \textbf{\# Parameters}  \\
\hline
MulT (previous SOTA)    &  2.24 M \\
\hline
\CoolName (ours)    &  \textbf{0.14 M} \\
\hline
\end{tabular}
\caption{Number of model parameters (M = Million).}
\label{table:params}
\end{table}

\subsection{Baselines}
For basleine evaluations, we use Early Fusion LSTM (EF-LSTM) and Late Fusion LSTM (LF-LSTM) \citep{tsai-etal-2019-multimodal} as baselines. In addition, we compare our model against similar methods as in previous works \cite{tsai-etal-2019-multimodal}, which combine a Connectionist Temporal Classification (CTC) loss \cite{graves2006connectionist} with the pre-existing methods such as EF-LSTM, MCTN \cite{pham2019found}, RAVEN \cite{wang2019words}.

\subsection{Results}
Shown in Table \ref{table:iemocap} and Table \ref{table:mosi}, \CoolName substantially out-performs previous methods on unaligned IEMOCAP benchmark and CMU-MOSI benchmark on most of the metrics. \CoolName also achieves on-par performance on the $\text{Acc}_7$ metric on CMU-MOSI benchmark. With an extremely small number of parameters, our model is able to learn better alignment and fusion for multimodal sentiment analysis task. Details regarding our model and hyper-parameters can be found in the Appendix \ref{sec:appendix}.

\paragraph{Parameter Efficiency (\CoolName vs MulT)}
We discover that \CoolName is a highly parameter-efficient model. A comparison of model parameters between \CoolName and MulT \cite{tsai-etal-2019-multimodal} (previous state of the art) is shown in Table \ref{table:params}. The hyperparameter used for this comparison can be found in the Appendix. With only a fraction ($6.25\%$) of MulT's parameters, \CoolName is able to achieve on-par, and in most cases superior performance on the two datasets. This demonstrates the parameter efficiency of our method.

\paragraph{Qualitative Analysis}
The attention weights on the graph edges forms a natural way to interpret our model. We visualize the edges to probe what \CoolName has learned. The following case study is a randomly selected video clip from the CMU-MOSI validation set. We observe the phenomena shown below is a general trend.

\begin{figure*}[t]
    \centering
    \begin{subfigure}{0.4\textwidth}
        \centering
        \includegraphics[width=\linewidth]{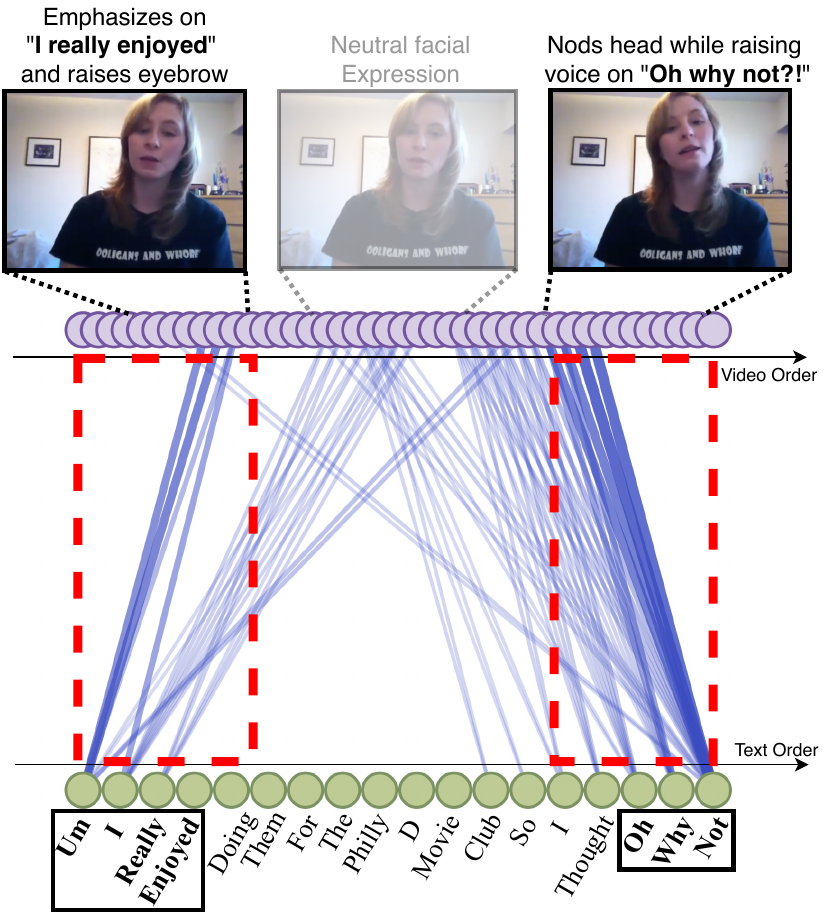}
        \caption{Text-to-vision edge attention weights. }
        \label{fig:bi-modal-vis-a}
    \end{subfigure}
    \begin{subfigure}{0.4\textwidth}
        \centering
        \includegraphics[width=\linewidth]{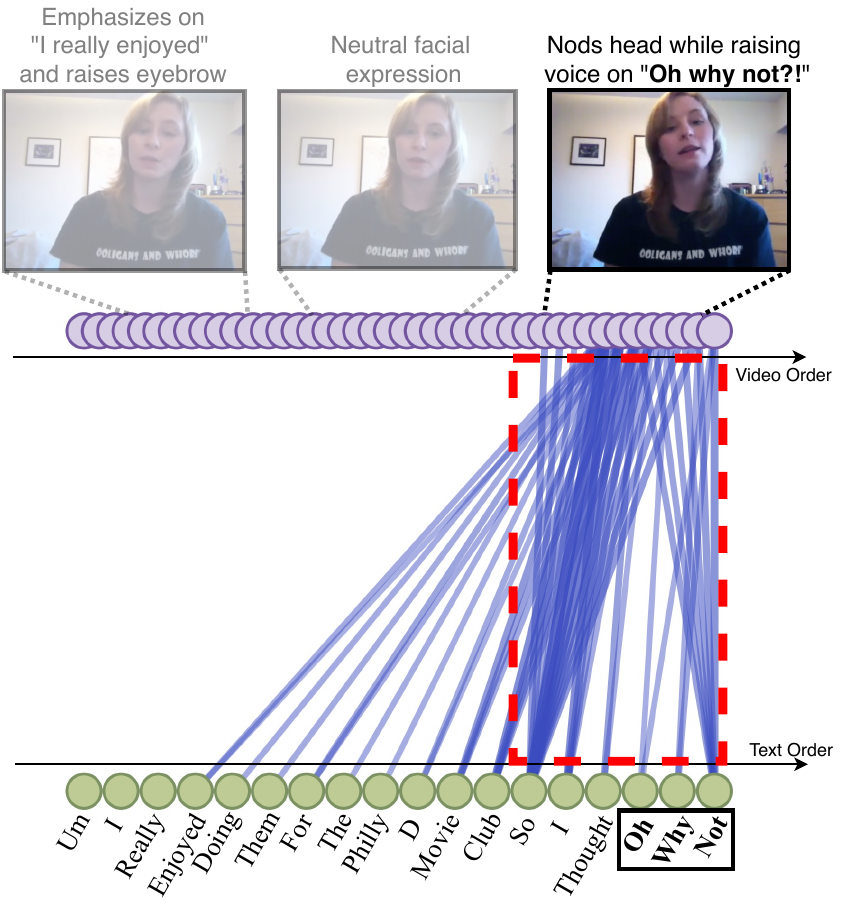}
        \caption{Vision-to-text edge attention weights. }
        \label{fig:bi-modal-vis-b}
    \end{subfigure}
    
    \caption{We visualized the asymmetric attention weights of the text-to-vision and vision-to-text edges for one of the validation sequence in CMU-MOSI dataset. The visualized attention weights are from layer 3 of \CoolName. Note that the edge types are not shown here for visual clarity.}
    \label{fig:bi-modal-vis}
\end{figure*}

In Figure \ref{fig:bi-modal-vis}, we show an example of the asymmetric bi-modal relations between vision and text. We observe that our model picks on meaningful relations between words such as ``\textbf{I really enjoyed}" and facial expressions such as raising eyebrow, highlighted in the red dashed boxes in Figure \ref{fig:bi-modal-vis-a}. Our model can also learn long-range correlation between ``\textbf{I really enjoyed}" and head nodding. Interestingly, we discover that strong relations that are not detected by vision-to-text edges can be recovered by the text-to-vision edges. This advocates the design of the multi-type edges, which allows the model to learn different relations independently that can complement one another.

Figure \ref{fig:tri-modal-vis} gives a holistic view of the attention weights among all three modalities. We observe a pattern where almost all edges involve the text modality. A possible explanation for this observation is that the text is the dominant modality with respect to the sentiment analysis task. This hypothesis is verified by the ablation study in Sec. \ref{sec:ablation_modality}. Meanwhile, there appears to be very small amount of edges connecting directly between vision and audio, indicating that there might be little meaningful correlation between them. This resonates with our ablation studies in Table \ref{table:ablation}, where vision and audio combined produce the lowest bi-modal performance. Under such circumstance, our \CoolName learns to kill direct audio-vision relations and instead fuse their information indirectly using the text modality as a proxy, whereas previous methods such as MulT keeps audio-vision attentions alive along the way, introducing possible spurious relations that could distract model learning.

\section{Ablation Study}

We conduct ablation study using unalgined CMU-MOSI dataset. \CoolName Full Model implements multimodal temporal edge types, adopts TopK edge pruning that keeps edges with top $80\%$ edge weights, and includes all three modalities as its input. Table \ref{table:ablation} shows the performance. We present research questions (RQs) as follows and discuss how ablation studies address them.

\subsection{RQ1: Does using 27 edge types help?}
We first study the effect of edge types on our model performance. As we incrementally add in multimodal and temporal edge types, our model's performance continues to increase. The model with 27 edge types performs the best under all metrics. By dedicating one attention vector $\mathbf{a}^{\phi_{ji}, \tau_{ji}}$ to each edge, \CoolName can model each complex relation individually, without having one relation interfering another. As shown in Figure \ref{fig:bi-modal-vis} and Table \ref{table:ablation}, such design enhances multimodal fusion and alignment, helps maintain long-range dependencies in multimodal sequences, and yields better results.
\begin{table}[t]
\renewcommand\arraystretch{1.3}
\centering
\resizebox{1.0\linewidth}{!}{
\begin{tabular}{c c c c} 
\toprule
\textbf{Ablation}  &  \textbf{$\text{Acc}_2\uparrow$} & \textbf{$\text{F1}\uparrow$} & \textbf{$\text{MAE}\downarrow$} \\ 
\toprule
\multicolumn{4}{c}{\textbf{Edge Types}} \\ 
\midrule
No Edge Types       &   82.4    &   82.5   &   0.937   \\
Multimodal Edges Only       &   85.6    &   85.7    &   \bf 0.859   \\
Temporal Edges Only       &   85.2    &   85.2    &   0.887   \\
\midrule
\multicolumn{4}{c}{\textbf{Pruning}} \\
\midrule
Random Pruning Keep $80\%$  &   75.5 &  74.5 & 1.080  \\
No Pruning                  &   84.7 &  84.7 & 0.908  \\
\midrule
\multicolumn{4}{c}{\textbf{Modalities}} \\
\midrule
Language Only       &   81.5 & 81.4 & 0.911 \\
Vision Only         &   57.0 & 57.1 & 1.41 \\
Audio Only          &   58.1 & 58.1 & 1.37 \\

Vision, Audio       &   62.0 &  59.2 & 1.360  \\
Language, Audio     &   85.9 &  85.7 & 0.915  \\
Language, Vision    &   86.6 &  86.6 & 0.896  \\
\midrule
Full Model, All Modalities    &   \bf 87.0 & \bf 87.0 & \bf 0.859   \\ 
\bottomrule
\end{tabular}
}
\caption{Ablation on unaligned CMU-MOSI validation set.}
\label{table:ablation}
\end{table}
\subsection{RQ2: Does our pruning method help?}
\label{sec:albation_pruning}
We compare our TopK edge pruning to no pruning and random pruning to demonstrates it effectiveness. We find that TopK pruning exceeds both no pruning and random pruning models in every aspect. It is clear that, by selectively keeping the top 80\% most important edges, our model learns more meaningful representations than randomly keeping 80\%. Our model also beats the one where no pruning is applied, which attests to our assumption and observation from previous work \citep{Lee2019SelfAttentionGP, knyazev2019understanding} that spurious correlations do exist and can distract model from focusing on important interactions. Therefore, by pruning away the spurious relations, the model learned a better representation of the interactions, while using significantly fewer computation resources.

\subsection{RQ3: Are all modalities helpful?}
\label{sec:ablation_modality}
Lastly, we study the impact of different modality combinations used in our model. As shown in Table \ref{table:ablation}, we find that adding a modality consistently brings performance gains to our model. Through the addition of individual modalities, we find that adding the text modality gives the most significant performance gain, indicating that text may be the most dominant modality for our task. This can also be qualitative confirmed by seeing the concentrated edge weights around text modality in Figure \ref{fig:tri-modal-vis}. This observation also conforms with the observations seen in prior works \cite{tsai-etal-2019-multimodal, pham2019found}. On the contrary, adding audio only brings marginal performance gain. Overall, this ablation study demonstrates that all modalities are beneficial for our model to learn better multimodal representations.
\section{Conclusion}

In this paper, we presented the Modal-Temporal Attention Graph (\CoolName). We showed that \CoolName is an interpretable model that is capable of both fusion and alignment. It achieves similar to SOTA performance on two publicly available datasets for emotion recognition and sentiment analysis while utilizing substantially lower number of parameters than a transformer-based model such as MulT.

\section*{Acknowledgements}
We thank Jianing Qian, Xiaochuang Han and Haoping Bai at CMU and the anonymous reviewers at NAACL for providing helpful discussions and feedbacks. This material is based upon work partially supported by BMW of North America, the National Science Foundation and National Institutes of Health. Any opinions, findings, and conclusions or recommendations expressed in this material are those of the authors and do not necessarily reflect the views of BMW of North America, National Science Foundation or National Institutes of Health, and no official endorsement should be inferred.

\bibliography{custom}
\bibliographystyle{acl_natbib}

\clearpage
\appendix

\section{Appendix}
\label{sec:appendix}

\subsection{Positional Embedding}

\begin{align}
    PE_{(pos,2i)} &= sin(pos / 10000^{2i/d_{emb}}) \\
    PE_{(pos,2i+1)} &= cos(pos / 10000^{2i/d_{emb}})
\end{align}

\subsection{Details regarding pseudo-alignment \label{sec:appendix_pseudo_align}}

\begin{figure}[h]
    \centering
    \begin{subfigure}{0.9\linewidth}
        \includegraphics[width=\linewidth]{figs/pdf/psudo-align.pdf}
        \caption{Pseudo-Alignment example with less vision nodes}
    \end{subfigure}
    \begin{subfigure}{0.9\linewidth}
        \includegraphics[width=\linewidth]{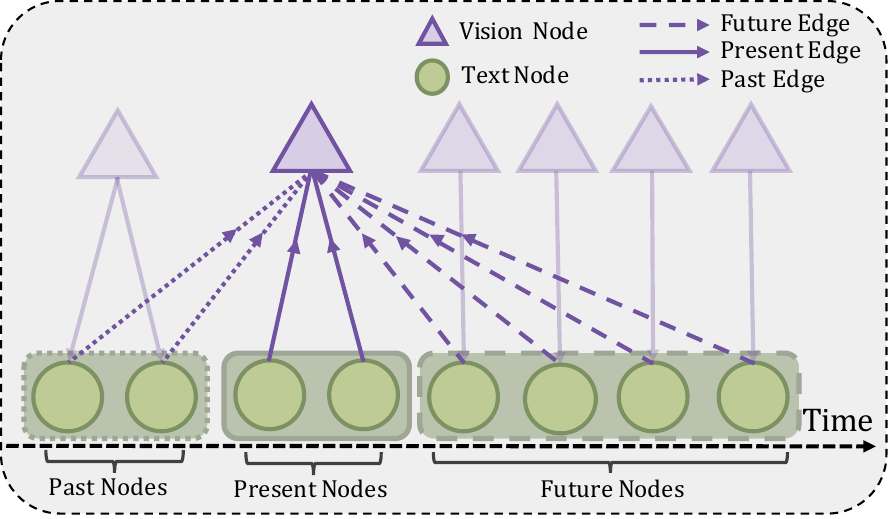}
        \caption{Pseudo-Alignment example with more vision nodes}
    \end{subfigure}
    \caption{Examples of pseudo-align heuristic to coarsely define past, present and future relationships between two unaligned modalities. We try to spread and match the two modalities as much as uniformly possible (the top figure). When the shorter modality contains more and more nodes, we align as many nodes from the shorter sequence as possible with a minimum alignment window size of 2 to the longer sequence, and the rest nodes from the shorter sequence are aligned with window size of 1 (the bottom figure).}
    \label{fig:pseudo-align}
\end{figure}
For node $v_i$ in $\pi_i$, in order to determine the ``present" nodes $\mathbfcal{V}_{present}$ in a different modality, we draw an analogy from 1D convolution operation. We are given two sequences of different lengths, and we can treat the longer sequence as input and shorter sequence as output to a Conv1D operation. Our goal is to find a feasible stride and kernel size that aligns the input and output. The kernel size defines how many nodes from the longer sequence to be aligned as ``present" to each node from the shorter sequence.  The stride size defines how far away such alignments should spread in time. We do not consider any padding and have the following equation in Conv1D operation:

\begin{equation}
    \frac{M - W}{S} + 1 = N \label{eq:conv1d}
\end{equation}
where $M$ and $N$ are the sequence lengths of the output and input to a Conv1D operator, respectively. $W$ is the kernel size and $S$ the stride size. From Eq. \ref{eq:conv1d}, we can further write the relationship between $W$ and $S$ as $W = M - (N - 1) * S$. It is clear that the minimum stride size is 1 to a Conv1D operation, and the maximum is $\lfloor \frac{M}{N-1} \rfloor$ in order to keep $W$ positive. We take the average of the minimum and maximum possible values of $S$ as our stride size. In case that $N > \frac{M}{2}$, we set window size as 2 and stride as 2. We then find the maximal number of nodes from $N$ that can have kernel size of 2, and the rest of the nodes will have kernel size of 1. Eq. \ref{eq:stride_size} shows our kernel size and stride size calculation and Figure. \ref{fig:pseudo-align} illustrates our pseudo-alignment heuristic.

\begin{equation}
\begin{cases}
    S = \lceil \textbf{avg}(1, \lfloor \frac{M}{N-1} \rfloor) \rceil, & \multirow{2}{*}{\textbf{if    } $N \leq \frac{M}{2}$} \\ 
    W = M - (N - 1) * S & \\
    S = 2, W = 2 & \textbf{otherwise}
\end{cases}
\label{eq:stride_size}
\end{equation}

\subsection{Model Efficiency}

\begin{figure}[t]
    \centering
    \includegraphics[width=\linewidth]{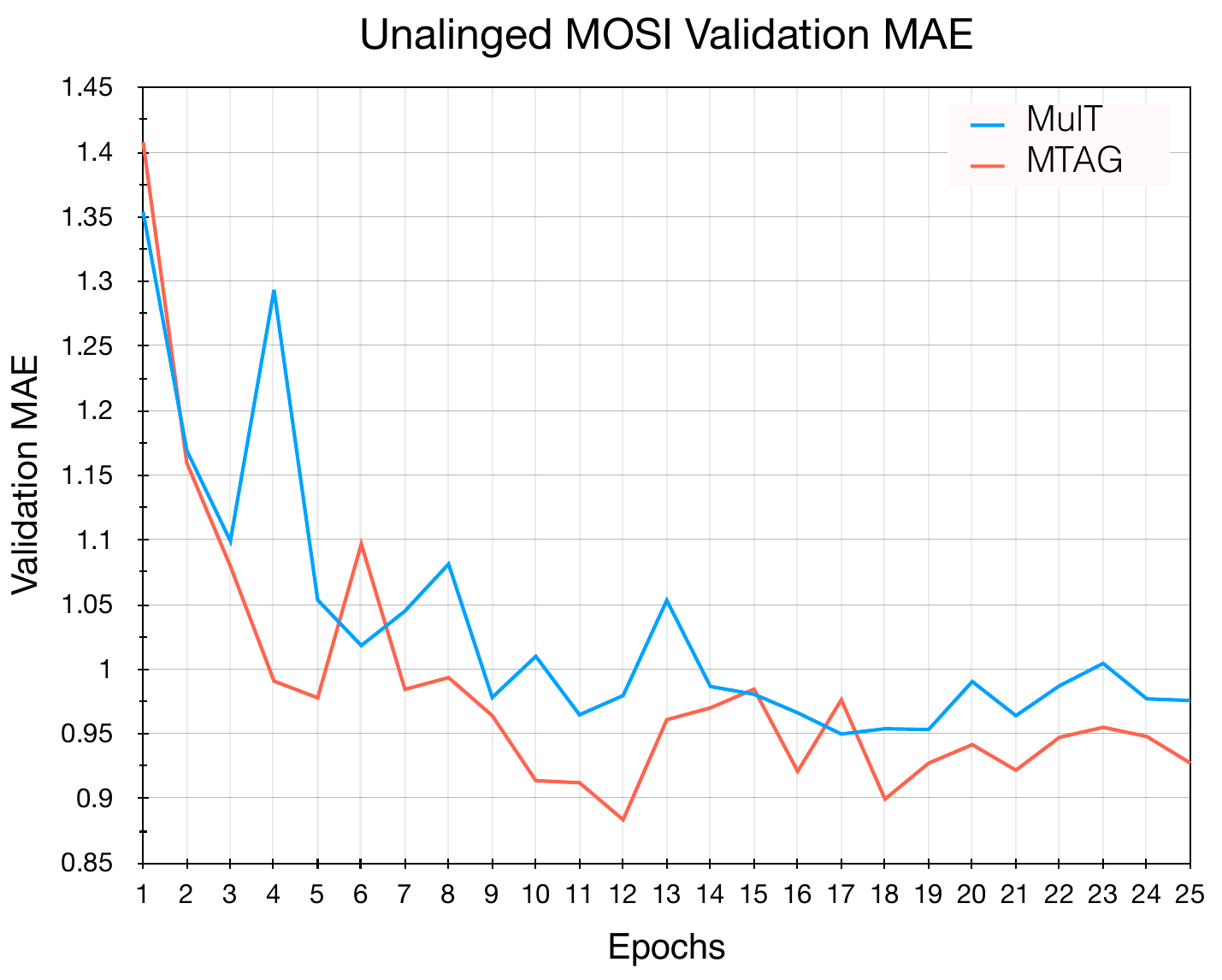}
    \caption{Convergence comparison between \CoolName and MulT on validation set of unaligned CMU-MOSI dataset.}
    \label{fig:convergence}
\end{figure}

\begin{table*}[t]
\renewcommand\arraystretch{1.3}
\centering
\resizebox{0.6\textwidth}{!}{
\begin{tabular}{c c c} 
\toprule
\textbf{Hyperparameter}  &  \textbf{CMU-MOSI} & \textbf{IEMOCAP} \\ 
\toprule
Batch Size                      &   64      & 64   \\
Initial Learning Rate           &   1e-3    &  1e-3  \\
Optimizer                       &   Adam    &   Adam \\
Number of \CoolName Layers          &   6       &  2  \\
Number of Attention Heads &   4       &   8 \\
Node Embedding Dimension          &   64      & 64   \\
Edge Pruning Keep Percentage      &   80$\%$  & 80\%   \\
Total Epochs                    &   20     &  35  \\
\bottomrule
\end{tabular}
}
\caption{Hyperparameter settings on CMU-MOSI and IEMOCAP benchmarks.}
\label{table:hp}
\end{table*}
\paragraph{Number of Parameters.} We compare the parameter efficiency of our model against the SOTA model, the Multimodal Transformer (MulT) \cite{tsai-etal-2019-multimodal}. We first look at the total number of parameters used by the two models. Table \ref{table:params} illustrates that our model uses 0.14 million parameters, only $6.3\%$ of those in MulT, which has 2.24 million parameters, and yet still achieves state-of-the-art performance. We attribute this result to the effective early fusion of multiple (more than 2) modalities using the \CoolName component. In MulT, trimodal fusion happens at a very late stage of the architecture, since each cross-modal Transformer can only model bi-modal interactions. This late fusion regime requires earlier layers to preserve more original information, and thus resulting in a need for more model parameters. 

\begin{table}
\centering
\large
\renewcommand\arraystretch{1.4}
\resizebox{\linewidth}{!}{\begin{tabular}{c |c |c }
\hline
\textbf{Model} \textbackslash \ \textbf{Dataset} & \textbf{CMU-MOSI}  & \textbf{IEMOCAP}  \\ 
\hline
MulT             &  $27.2_{\pm 2.33}$           &     $56.0_{\pm 4.59}$               \\ 
\hline
MTAG (ours)     &  $\bf 24.4_{\pm 0.95}$       & $ \bf 44.4_{\pm 0.55}$  \\
\hline
\end{tabular}}
\caption{Training time per epoch (in seconds) comparison. Run time is averaged over 5 training epochs, with the subscript denoting standard deviation. Both models use batch\_size=32, num\_layers=6 and are run on the same machine with a single GPU of Nvidia GTX 1080 Ti. Both benchmarks used are the unaligned version.}
\label{table:time}
\end{table}

\paragraph{Convergence.} Figure \ref{fig:convergence} gives a comparison between the convergence speed of our model and MulT. Both models are trained with batch\_size=32 and lr=1e-3, with the default (best) hyperparameters used for both models. We use the unalignd CMU-MOSI for this study. We observe \CoolName converges much faster at epoch 12, comparing with MulT at epoch 17. In addition, we see that our validation MAE curve on the unaligned MOSI validation set goes consistently below MulT's curve. This faster convergence performance could be explained by the small amount of parameters \CoolName uses - \CoolName has a much smaller parameter search space for the optimizer, resulting in faster training and earlier convergence.  

\paragraph{Training time.} We also compare how fast our model runs against MulT. Specifically, under the same condition, we calculate the time it takes for each model to run training for 1 epoch. Table \ref{table:time} shows the details. We can see that our model runs significantly faster than MulT on both benchmarks, which can be attributed to our light-weight model design (as shown in Table \ref{table:params_detail}). Meanwhile, our edge pruning also reduces the number of computation by throwing away edges that are deemed less important by the model, thus improving the run-time of our model.

\paragraph{Overall Efficiency.} From the perspective of training time, number of parameters used, and convergence analysis, it is clear that our model is capable of achieving better results while using much smaller amount of computational resources than the previous state of the art.

\begin{table}[t]
\centering
\renewcommand\arraystretch{1.4}
\begin{tabular}{ c | c}
\hline
\textbf{Model} & \textbf{\# Parameters}  \\
\hline
MulT (previous SOTA)    &  2,240,921 \\
\hline
MTAG (ours)            & 142,363 \\
\hline
\end{tabular}
\caption{Number of model parameters comparison.}
\label{table:params_detail}
\end{table}



\subsection{Hyperparameters}
We elaborate on the technical details including hyperparameter settings in Table \ref{table:hp}. We conduct a basic grid search to find good hyperparameters such as initial learning rate, number of \CoolName layers etc. We use Adam as our optimizer and decays the learning rate by half whenever the validation loss plateaus. Notice that we are using a design that roughly yields a model with a similar structure as in previous works such as MulT. Nevertheless, we still manage to use far less number of parameters during optimization. We use one NVIDIA GTX 1080 Ti for training and evaluation. In addition, the model and hyperparameters we use for ablation study are the same as the ones used for the main experiment, both of which are conducted on CMU-MOSI.

\subsection{Number of Parameters Comparison}
For a fair comparison on number of parameters between \CoolName and MulT, we use the same number of layers and attention heads for both models (i.e. 6 layers of MulT with 4 attention heads). A detailed comparison is shown in Table \ref{table:params_detail}.

\end{document}